\documentclass[a4paper,conference]{IEEEtran}
\IEEEoverridecommandlockouts

\usepackage{cite}
\usepackage{amsmath,amssymb,amsfonts}
\usepackage{algpseudocode}
\usepackage{graphicx}
\usepackage{pdflscape}
\usepackage{textcomp}
\usepackage{xcolor}
\usepackage{algorithm}
\usepackage[font=footnotesize,skip=2pt]{caption}

\def\BibTeX{{\rm B\kern-.05em{\sc i\kern-.025em b}\kern-.08em
    T\kern-.1667em\lower.7ex\hbox{E}\kern-.125emX}}

\title{AI-Driven Real-Time Relay Optimisation in Smart Urban NR-V2X Networks via Learning-to-Optimise Graph Neural Networks

}

\author{
\IEEEauthorblockN{Giambattista Amati, Federica Mangiatordi, Emiliano Pallotti, Simone Angelini}
\IEEEauthorblockA{\textit{Fondazione Ugo Bordoni}\\
Rome, Italy\\
Emails: \{gamati, fmangiatordi, epallotti,sangelini\}@fub.it}
}

\begin{document}
\maketitle

\begin{abstract}
 Reliable and low-latency communication is a fundamental requirement for smart city services and Industry 4.0 applications enabled by NR-V2X networks. However, limited Road-Side Unit (RSU) deployment and complex urban propagation conditions often prevent Connected and Automated Vehicles (CAVs) from maintaining stable connectivity.
This paper proposes an AI-driven Learning-to-Optimise (L2O) framework based on Graph Neural Networks (GNNs) for real-time multi-hop relay selection in NR-V2X systems. The vehicular network is modelled as a graph, where nodes represent CAVs and RSUs, and edges encode radio-link characteristics. An offline Mixed-Integer Linear Programming (MILP) formulation provides optimal relay decisions used as supervision for training an edge-aware Graph Isomorphism Network with Edge Features (GINE).
Extensive experiments on realistic urban datasets demonstrate that the proposed approach achieves near-optimal connectivity performance, recovering up to 11.3\% connectivity gain, while reducing execution time by orders of magnitude (up to 100× speed-up) compared to MILP. The framework enables scalable and real-time network control, making it suitable for smart city and Industry 4.0 deployments.

\end{abstract}


\section{Introduction}

The evolution of Industry 4.0 and smart city infrastructures is driving the need for communication systems that are not only reliable and low-latency, but also intelligent, adaptive, and aware of their operating environment. In cyber-physical domains such as connected mobility, communication decisions must continuously adapt to variations in topology, traffic demand, and radio conditions.

NR-V2X is a key enabling technology for advanced mobility services, including cooperative perception, automated driving, and real-time traffic coordination. Modern Connected and Automated Vehicles (CAVs) integrate heterogeneous onboard sensing platforms, such as LiDAR, radar, and cameras, which generate high-rate data streams that require robust uplink connectivity toward roadside or edge infrastructure \cite{ahangar2021survey,wang2018networking}. However, in dense urban environments, the reliability of direct vehicle-to-infrastructure communication is often degraded by non-line-of-sight propagation, blockage, limited Road-Side Unit (RSU) density, and fast topology changes \cite{Sun2018PropagationMA,9119160}.

Multi-hop relay-assisted communication can mitigate these limitations by exploiting neighbouring vehicles as intermediate forwarding nodes, thus extending infrastructure reach without requiring additional fixed deployments \cite{10304089}. Nevertheless, relay selection under realistic radio, capacity, and routing constraints leads to a combinatorial optimisation problem that is typically solved via Mixed-Integer Linear Programming (MILP). While MILP provides optimal relay configurations, its runtime grows rapidly with graph size and density, making it unsuitable for real-time deployment in dense urban scenarios.

To address this limitation, this paper proposes a deep learning-based Learning-to-Optimise (L2O) framework for real-time relay selection in NR-V2X. 

More importantly, this work introduces a paradigm shift by 
reformulating combinatorial optimisation problems as context-aware 
learning tasks over structured environments. This perspective enables 
scalable real-time decision-making in scenarios where traditional 
optimisation becomes computationally prohibitive. To the best of our 
knowledge, this is the first work that unifies the context-aware system 
modelling, graph-based representation, and Learning-to-Optimise for 
real-time multi-hop relay selection under realistic NR-V2X constraints.

In this perspective, relay selection is no longer treated as a standalone optimisation problem but as a context-reasoning process over graph-structured data. 
The system continuously acquires contextual information from the communication environment, encodes it as a graph, processes it through a Graph Neural Network, and adapts relay decisions in real time.
Unlike traditional context-aware communication mechanisms based on local rules or handcrafted policies, the proposed framework integrates context representation, context reasoning, and decision adaptation into a unified learning-based architecture. The overall system architecture is illustrated in Fig.~1, 
highlighting the interaction between context acquisition, 
graph modelling, GNN-based reasoning, and relay decision making.

The main contributions of this paper are threefold:

\begin{itemize}
\item \textbf{Artificial Intelligence}: we introduce an AI-native 
deep learning framework based on Graph Neural Networks for learning 
near-optimal relay selection policies;

\item \textbf{Context-Aware Pervasive Systems}: we model NR-V2X networks 
as context-aware systems, where topology, radio conditions, and traffic demand 
are jointly exploited to enable adaptive decision-making;

\item \textbf{Smart Communications }: we design a real-time 
relay optimisation mechanism for vehicular networks, supporting smart city and cyber-physical system applications with stringent latency constraints.
\end{itemize}


\section{Related Work}

Early learning-based approaches for vehicular communication optimisation 
mainly relied on fully connected neural architectures trained to emulate 
conventional resource allocation or interference management algorithms 
\cite{8891446,8943940}. While effective in simplified settings, these 
methods typically flatten the network state into tabular features and 
do not explicitly model the relational structure of vehicular communication 
graphs. As a result, they exhibit limited scalability and poor generalisation 
in dense and highly dynamic environments.

Graph Neural Networks (GNNs) have recently emerged as a more suitable 
paradigm for wireless and vehicular networking problems, since vehicles, 
RSUs, and wireless links can be naturally represented as graph entities. 
Existing GNN-based studies have addressed tasks such as power control, 
spectrum management, scheduling, and connectivity prediction 
\cite{9322537,ji2025graphneuralnetworksdeep,amati2025AEIT,HUAN20255427}. 
More broadly, GNNs have been recognised as an effective tool for learning 
over relational data and structured environments \cite{wu2020comprehensive}. 
However, most existing works focus on single-hop or partial decision problems 
and do not address end-to-end multi-hop relay optimisation under global 
constraints, such as flow conservation, capacity limits, and loop avoidance.

A related line of work has explored edge-centric graph representations and 
link-level decision models, including line-graph transformations and 
co-embedding schemes for nodes and edges 
\cite{9979700,naderializadeh2021wirelesslinkschedulinggraph,11268640,10068338,
jiang2020coembeddingnodesedgesgraph,Amati2024AEIT}. While these approaches 
better capture link attributes, they often introduce additional computational 
complexity and still lack a unified framework for real-time relay activation 
under global communication constraints.

In parallel, context-aware and pervasive systems have emphasised the role of 
environmental information in enabling adaptive system behaviour. However, in 
communication networks, context-awareness has traditionally been implemented 
through rule-based or locally optimised mechanisms, rather than through 
learning-based reasoning over structured relational data \cite{8784545}.

More recently, Learning-to-Optimise (L2O) approaches have been proposed to 
approximate solutions of complex optimisation problems using machine learning 
models \cite{BENGIO2021405}. These methods aim to replace or accelerate 
computationally expensive solvers, but their application to structured communication problems with strict global constraints remains limited.

This paper advances the state of the art by bridging these directions within 
a unified framework. Specifically, we propose a graph-based Learning-to-Optimise 
approach in which: i) an offline MILP oracle captures global optimality under 
NR-V2X constraints; ii) the environmental context is encoded as an attribute 
communication graph; and iii) an edge-aware GNN performs context reasoning 
and real-time relay adaptation.

Despite recent progress, no existing approach jointly captures 
global optimisation constraints, structured relational modelling, 
and real-time decision-making in dynamic NR-V2X environments.

\section{Context-Aware System Modeling}

\begin{figure*}[h!]
\centering
\includegraphics[width=0.8\textwidth]{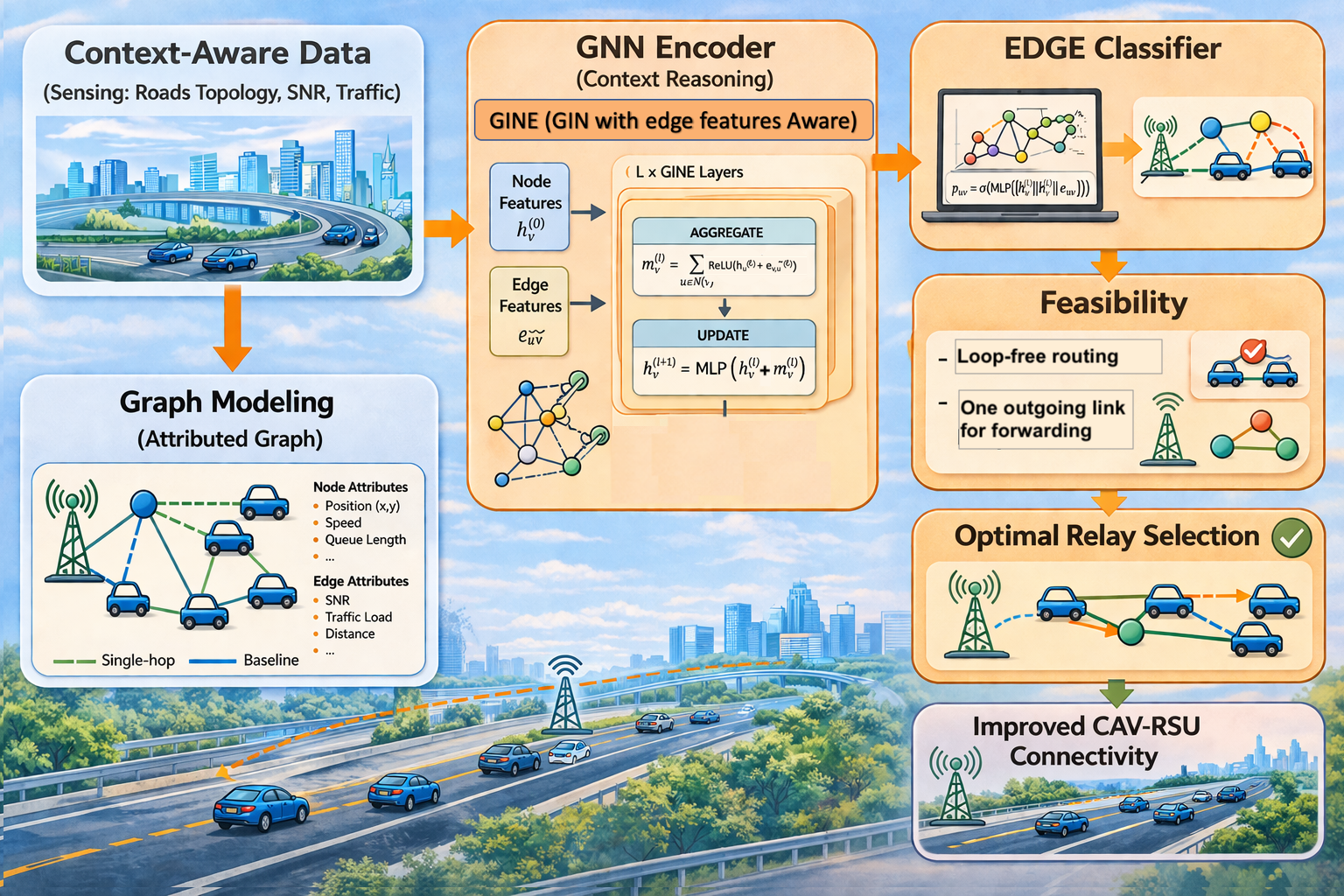}
\caption{\textbf{AI-native context-aware pipeline.} Environmental context is acquired, 
encoded as an attributed graph, processed by a GNN-based reasoning 
module within a Learning-to-Optimise framework, and mapped into 
relay selection decisions with feasibility refinement under 
communication constraints.}
\label{fig:context_gnn}
\end{figure*}

The proposed framework can be rigorously interpreted as a context-aware 
pervasive system, in which communication decisions are driven by the 
continuous acquisition, representation, and processing of environmental 
information. As illustrated in Fig.~\ref{fig:context_gnn}, the proposed framework follows a 
structured pipeline integrating sensing, graph modelling, 
learning-based reasoning and decision-making.

In the considered NR-V2X scenario, context is defined as the set of 
variables that directly influence communication performance and 
relay feasibility.

\subsection{Context Definition}

The relevant context includes:
\begin{itemize}
    \item the instantaneous communication topology, determined by the spatial distribution of CAVs and RSUs;
    \item radio-layer conditions, including SNR, SINR, and achievable link capacity;
    \item the traffic demand generated by vehicles;
    \item spatial and neighbourhood relationships among nodes.
\end{itemize}

These contextual factors evolve continuously over time due to mobility, 
channel variability and dynamic traffic patterns, requiring adaptive 
decision-making mechanisms.

\subsection{Context Representation}

Rather than treating these variables independently, the proposed method 
encodes context as an attributed directed graph. Nodes represent CAVs 
and RSUs, while edges correspond to feasible communication links enriched 
with radio-aware features. This representation provides a unified and structured description 
of the system state, capturing both relational dependencies and 
heterogeneous attributes within a single model.

In this work, radio context is generated through the 
OpenStreetMap--SUMO--GEMV$^2$/Sionna simulation pipeline 
and represents the simulated counterpart of PHY-layer 
link-quality estimates available in real CAV systems.
\subsection{Context Reasoning and Adaptation}

Within this framework, the GNN operates as a context reasoning engine, 
mapping structured environmental information into relay activation decisions. 
Through message passing, the model learns high-level representations that 
capture both local interactions and global multi-hop dependencies across 
the network.
Based on these learned representations, the system dynamically selects 
relay links, enabling adaptive communication strategies under varying 
topological and radio conditions. This process allows the model to 
implicitly approximate the underlying optimisation problem while 
maintaining real-time execution.
Different from conventional context-aware systems based on rule-based adaptation or local heuristics, the proposed approach performs end-to-end 
context learning, where context representation and decision-making are 
jointly optimised within a unified deep learning architecture.

This perspective is central to the contribution of the paper: the proposed framework is not merely a predictor of relay links, but an AI-native context-aware communication system that enables real-time adaptation in 
dynamic cyber-physical environments.

\section{Graph Representation}

Each V2X snapshot corresponds to the instantaneous communication topology of the vehicular system and is modelled as a directed graph \begin{equation}
G_t=(V_t,E_t).
\end{equation} The node set includes both CAVs and RSUs, i.e., $V_t = V_{\text{CAV}} \cup V_{\text{RSU}}$, while directed edges $(i,j)\in E_t$ represent feasible wireless links satisfying the NR-V2X SNR threshold.
Nodes are characterised by spatial coordinates, a CAV/RSU role indicator, and, for CAVs, the current uplink traffic demand associated with 3GPP NR-V2X services. Edges carry radio-aware features, including SNR, SINR, LoS/NLoS classification, link distance, and the Shannon capacity computed from the instantaneous SINR. As vehicles move and channel conditions change, the communication graph evolves accordingly. We model the complete network trace as a time-indexed sequence $\{G_t\}_{t=0}^{T-1}$ sampled every 1~s.

\subsection{Simulation Framework}
Vehicular mobility was generated using SUMO \cite{SUMO2,SUMO} 
on top of OpenStreetMap topologies \cite{OpenStreetMap}, 
representing four realistic urban areas of Rome, namely 
Porta Pia, Trastevere, Ostiense, and Marconi, which are 
characterized by heterogeneous traffic density and propagation conditions.
\begin{table}[h]
\centering
\caption{Characteristics of the considered urban areas}
\label{tab:areas}
\begin{tabular}{lccc}
\hline
\textbf{Area} & \textbf{Urban Type} & \textbf{Traffic Density} & \textbf{Propagation} \\
\hline
Porta Pia   & Urban grid        & Medium & Mixed LoS/NLoS \\
Trastevere  & Dense urban       & High   & NLoS-dominated \\
Ostiense    & Mixed urban       & Medium & Mixed \\
Marconi     & Arterial/urban    & High   & Mixed LoS/NLoS \\
\hline
\end{tabular}
\end{table}

 The SUMO-generated CAV positions were then fed into a hybrid GEMV$^2$/Sionna radio simulator \cite{GEMV2,sionna}, which models geometry-aware path loss, LoS/NLoS conditions, shadowing, and multipath effects, ensuring PHY-layer consistency with NR-V2X assumptions.
Radio simulations adopt the following parameters: RSU transmit power 10~dBm, vehicle transmit power 0~dBm, antenna gains 1~dBi, carrier frequency 5.9~GHz, and 800~MHz bandwidth. At each timestamp, the combined mobility--radio pipeline produces a complete V2X snapshot with physically grounded connectivity patterns. Although based on realistic PHY-aware simulations reproducing heterogeneous urban propagation conditions,
the current validation does not yet include field trials
or operational NR-V2X deployments.

\subsection{Dataset Generation}

The dataset consists of 520,000 labelled V2X graph instances sampled 
at 1 s intervals. Its size results from combining multiple urban areas, 
time snapshots, and heterogeneous traffic and load configurations. 
To ensure effective training, it is balanced across the four considered 
urban areas, providing a representative distribution of topological 
and propagation conditions.
Each snapshot is processed using the MILP formulation described 
in Section~\ref{sec:milp_oracle}, which serves as an oracle to compute 
the optimal multi-hop relay configuration. The resulting activation 
matrix is used as edge-level supervision.

The data span heterogeneous traffic conditions, from sparse to dense 
vehicular regimes, yielding graphs with varying numbers of nodes and 
candidate radio links. Samples are partitioned into training, validation, 
and test sets according to a 70\%--10\%--20\% split.
\section{MILP Oracle for Relay-Link Optimization}
\label{sec:milp_oracle}

The optimal relay configuration is obtained by solving a 
Mixed-Integer Linear Programming (MILP) problem that models 
radio, flow, and capacity constraints in NR-V2X multi-hop 
communications. The MILP formulation acts as an offline oracle, 
providing optimal link activations for each V2X snapshot and 
serving as supervision for the learning-based framework.

\subsection{Decision Variables}

To jointly model connectivity and traffic routing, we define:
\begin{itemize}
    \item $v_{ij}\in\{0,1\}$: binary variable indicating activation of link $(i,j)$;
    \item $f_{ij}\ge0$: flow routed over link $(i,j)$;
    \item $z_i\in\{0,1\}$: indicator that CAV $i$ is connected to the infrastructure;
    \item $u_i\in\{0,1,\dots,M-1\}$: ordering variable used for cycle elimination;
    \item $w_{ik}\in\{0,1\}$: auxiliary variable enforcing relay consistency.
\end{itemize}

\subsection{Objective}

The objective is to maximize the number of CAVs that establish 
a valid uplink path toward at least one RSU:
\begin{equation}
    \max \sum_{i\in\text{CAV}} z_i.
\end{equation}

\subsection{Key Constraints}

\noindent\textbf{Flow conservation:}
\begin{equation}
\sum_j f_{ij} - \sum_k f_{ki} = d_i z_i, 
\quad \forall i \in \text{CAV},
\end{equation}
where $d_i$ denotes the traffic demand of CAV $i$.

\noindent\textbf{RSU capacity:}
\begin{equation}
\sum_i f_{ij} \le M_{\text{RSU}_j}, 
\quad \forall j \in \text{RSU}.
\end{equation}

The available capacity at RSU $j$ depends on the current load 
of the serving cell and is modelled as:
\begin{equation}
M_{\text{RSU}_j} = (1 - \rho_j)\, C^{\text{max}}_{\text{RSU}},
\end{equation}
where $\rho_j \in [0,1)$ represents the load factor and 
$C^{\text{max}}_{\text{RSU}}$ is the maximum achievable capacity. 
In this work, $C^{\text{max}}_{\text{RSU}}$ is set to $1$~Gbps, 
while $\rho_j \in [0.3, 0.5, 0.7]$ models moderate to high network 
load conditions. This formulation captures the fraction of radio 
resources available for relay traffic under realistic operating conditions.

\noindent\textbf{Link capacity:}
\begin{equation}
f_{ij} \le C_{ij} \, v_{ij}.
\end{equation}

\noindent\textbf{Single outgoing link:}
\begin{equation}
\sum_{j : (i,j) \in \mathcal{E}_{\text{valid}}} v_{ij} = z_i,
\quad \forall i \in \text{CAV}.
\end{equation}

\noindent\textbf{Relay consistency:}
\begin{align}
    w_{ik} &\le v_{ik}, &
    w_{ik} &\le z_k, \\
    w_{ik} &\ge v_{ik} + z_k - 1, &
    z_i    &\ge w_{ik}.
\end{align}

\noindent\textbf{Cycle elimination:}
\begin{equation}
u_i + 1 \leq u_j + M \left(1 - v_{ij}\right),
\quad \forall (i,j) \in \mathcal{E}_{\text{CAV--CAV}},
\end{equation}
where $M = |V_{\text{CAV}}|$ provides a valid upper bound for 
cycle-elimination constraints.

\subsection{Oracle Output}

For each V2X snapshot, the MILP produces an optimal binary 
activation matrix $\mathbf{A}^\star = \{A_{ij}^\star\}$. 
This matrix is used as supervision for the proposed deep 
learning model, enabling it to learn near-optimal relay 
decisions without solving the MILP at runtime.

\section{Edge-Aware GNN for Context Reasoning and Relay Activation}
\label{sec:l2o_framework}

The proposed Learning-to-Optimise surrogate consists of two components: i) an edge-aware Graph Neural Network that computes context-aware node representations; ii) an edge classifier that predicts relay activation probabilities.

\subsection{Node and Edge Features}

Each V2X graph is enriched with node and edge attributes reflecting geometric, functional, and radio-layer properties. Node features encode spatial position, node type, and traffic demand:
\begin{equation}
x_v = [x_v,\, y_v,\, \text{CAV flag},\, d_v].
\end{equation}

Edge features encode instantaneous link quality, including Shannon capacity:
\begin{equation}
e_{uv} = C_{\text{Shannon}}(u,v)
       = B \log_2\!\left(1 + \mathrm{SINR}_{uv}\right).
\end{equation}

\subsection{GINE Encoder}

To explicitly account for link-level context, we adopt a Graph Isomorphism Network with Edge Features (GINE). Unlike topology-only GIN variants, GINE incorporates edge attributes directly into message passing, thus enabling the model to jointly reason over graph structure and radio conditions.

At layer $\ell$, node updates follow the standard edge-aware aggregation principle:
\begin{equation}
h_i^{(\ell+1)} =
\mathrm{MLP}^{(\ell)}
\!\left(
h_i^{(\ell)}
+
\sum_{j\in\mathcal{N}(i)}
\mathrm{ReLU}(h_j^{(\ell)}+\tilde e_{ij}^{(\ell)})
\right)
\end{equation}
where $\tilde{e}_{ij}^{(\ell)}$ denotes the learned projection of the raw edge features.

\subsection{Edge-Level Relay Activation}

For each directed edge $(u,v)\in E_t$, relay suitability is computed by combining the endpoint embeddings with the edge feature vector:
\begin{equation}
\hat{y}_{uv}
= \sigma\!\left(
    \mathrm{MLP}\big(
        [h_u \Vert h_v \Vert e_{uv}]
    \big)
\right).
\end{equation}

The resulting score $\hat{y}_{uv}\in[0,1]$ represents the probability that link $(u,v)$ should be activated.

\subsection{Training Procedure}

The model is trained in a supervised manner using the optimal 
relay activation matrices generated by the MILP oracle. The 
learning task is formulated as a binary edge classification 
problem.
Training is performed using a weighted binary cross-entropy 
loss to account for class imbalance between active and inactive 
edges. The model is optimised using Adam with a learning rate 
of $10^{-3}$ and a batch size of 64. Training is carried out 
for 200 epochs.

The GINE encoder consists of three message-passing layers, each with a hidden size of 256, followed by an edge classifier implemented as an MLP. 
To enhance generalisation, a dropout rate of 0.3 is employed.
During inference, edge probabilities are thresholded to obtain 
binary decisions, followed by a lightweight feasibility refinement 
step enforcing the single-outgoing-link constraint and cycle 
removal.

\subsection{Inference}

At runtime, relay decisions are obtained through a single forward pass followed by thresholding and a lightweight feasibility refinement stage that enforces the single-outgoing-link condition and removes cycles. This enables real-time adaptation with near-constant inference latency.

\section{Experimental Results}

This section evaluates the proposed framework in terms of 
(i) link-level agreement with the MILP oracle, 
(ii) computational latency, and 
(iii) system-level connectivity performance. 
The results demonstrate that the proposed Learning-to-Optimize (L2O) 
approach achieves near-optimal performance while enabling real-time execution.
\subsection{Link-Level Performance}

We first evaluate the capability of the proposed edge-aware GINE model 
to reproduce the relay activation decisions provided by the MILP oracle. 
This task is formulated as a binary edge classification problem, where 
each link is classified as active or inactive.

At the nominal operating threshold ($\tau = 0.5$), the model achieves 
an accuracy of 0.9560, precision of 0.9512, recall of 0.9661, and 
F1-score of 0.9586 on the validation set, indicating a strong 
agreement with the optimal MILP solutions.

\begin{figure}[t]
    \centering
    \includegraphics[width=0.95\linewidth]{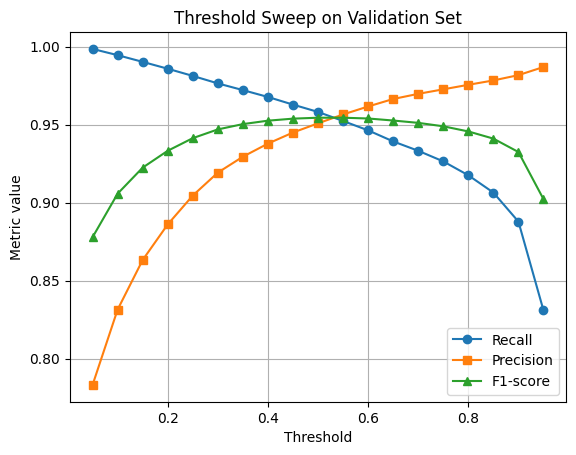}
    \caption{Impact of the decision threshold on link-level classification 
    performance. Precision and recall exhibit opposite trends, while the 
    F1-score remains stable over a wide operating range, indicating robust 
    behaviour with respect to threshold selection.}
    \label{fig:threshold_sweep}
\end{figure}

Fig.~\ref{fig:threshold_sweep} reports the impact of the decision threshold on the classification 
performance. As expected, recall decreases monotonically as the threshold 
increases, while precision exhibits the opposite trend. The F1-score shows 
a stable plateau in the range $[0.45, 0.60]$, indicating robust performance 
with respect to threshold selection.

At the standard operating point ($\tau = 0.5$), the model achieves a balanced 
trade-off between precision and recall. Lower thresholds enable 
recall-oriented configurations, with recall values approaching 1, which are 
particularly suitable for connectivity-preserving relay selection. Conversely, 
higher thresholds favour precision at the cost of reduced recall.
Overall, the proposed method attains performance close to the oracle, correctly identifying over 95\% of the MILP relay activations for $\tau = 0.5$, while substantially lowering the computational burden compared to exact optimisation.

\subsection{Runtime}

We next evaluate the computational efficiency of the proposed approach. 
While MILP exhibits a rapidly increasing runtime due to its combinatorial 
nature, the GNN-based model performs inference through a single forward 
pass, resulting in significantly lower and more predictable latency.

In all evaluated scenarios, inference time remains within a few milliseconds, 
enabling real-time relay adaptation in dynamic NR-V2X environments. 
Compared to MILP, the proposed framework achieves a reduction in execution time 
 of up to two orders of magnitude, while maintaining high solution quality.
\begin{figure}[h!]
    \centering
    \includegraphics[width=0.99\linewidth]{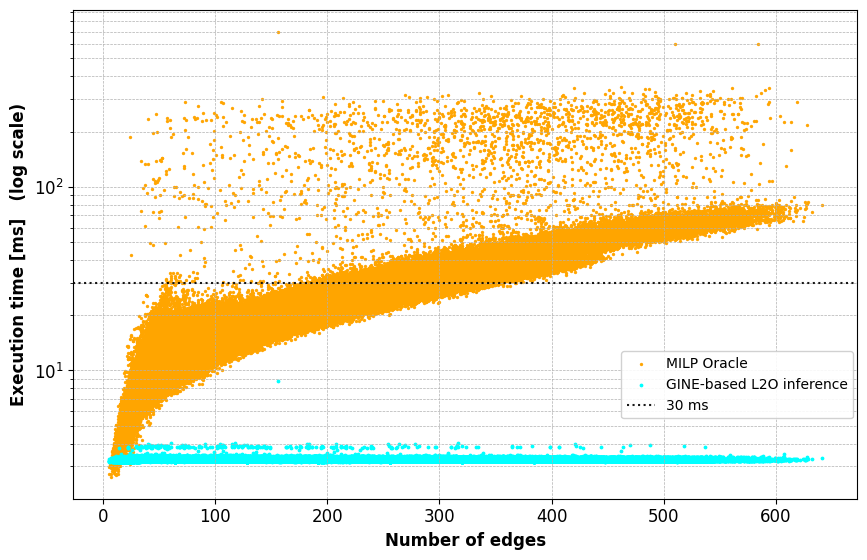}
    \caption{Execution time as a function of graph size for MILP and GNN-based inference. 
The proposed approach achieves near-constant latency, while MILP exhibits 
super-linear growth with graph complexity.}
    \label{fig:placeholder}
\end{figure}

These results confirm that the L2O approach provides a scalable and 
deployment-ready alternative to exact optimisation methods.

\subsection{Connectivity Gain}

Finally, we evaluate the impact of the proposed method on system-level 
connectivity, defined as the number of CAVs that successfully establish 
a multi-hop path toward the infrastructure.
Compared to single-hop communication baselines, multi-hop relay selection 
provides a substantial improvement in connectivity. However, the achievable 
connectivity gain is strongly influenced by both the infrastructure density 
and the network load conditions.
In the considered scenarios, the number of RSUs varies from 2 to 4 within 
a $1\,\text{km}^2$ urban area. Under sparse infrastructure configurations 
(e.g., 2 RSUs), multi-hop relaying increases the number of connected CAVs 
by approximately 8--11\% compared to single-hop communication. When the 
number of RSUs increases to 4, the relative gain reduces to about 6--8\%, 
as direct connectivity becomes more likely.
Connectivity performance is also affected by the RSU load state, denoted 
by $\rho$. For moderate load conditions ($\rho \approx 0.5$), the proposed 
approach maintains most of the connectivity gains, with improvements of 
around 8--11.3\%. Under high-load regimes ($\rho \geq 0.7$), the gain decreases 
to approximately 5--7\%, due to resource saturation at the infrastructure.

Across all evaluated scenarios, the proposed learning-based framework is able 
to recover more than 95\% of the connectivity achieved by the MILP oracle, 
while maintaining real-time execution.

These results demonstrate that AI-driven, context-aware relay selection 
effectively adapts to both infrastructure density and network load conditions, 
enabling scalable and efficient communication strategies for smart-city 
NR-V2X systems. Future work will investigate robustness under highly
dynamic and adversarial communication conditions,
as well as temporal graph extensions explicitly modelling
vehicular mobility evolution and sequential decision dynamics.

\section*{Acknowledgement} 
The work has been carried out in the framework of Spectrum Sharing project between the Ministry of Enterprises and Made in Italy (MIMIT) and Fondazione Ugo Bordoni.

\bibliographystyle{IEEEtran}
\bibliography{References.bib}
\newpage

\end{document}